\documentclass{article}
\usepackage{amsmath,amssymb} % For LaTeX2e
\usepackage{iclr2027_conference,times}

\usepackage{amsmath,amsfonts,bm}

\def\eqref#1{equation~\ref{#1}}
\def\1{\bm{1}}

\DeclareMathAlphabet{\mathsfit}{\encodingdefault}{\sfdefault}{m}{sl}
\SetMathAlphabet{\mathsfit}{bold}{\encodingdefault}{\sfdefault}{bx}{n}

\usepackage{hyperref}
\usepackage{url}
\usepackage{tabularx}
\usepackage{booktabs}
\usepackage{amssymb}
\usepackage{graphicx}

\title{AI Harness: Certification under Proposal-Conditioned Information for Foundation-Model Agents}

\author{
Hailin Zhong \\
Hong Kong Baptist University \\
Hong Kong, China \\
\And
Shengxin Zhu\thanks{Corresponding author.} \\
Beijing Normal University \\
Zhuhai, China \\
\texttt{shengxin.zhu@bnu.edu.cn}
}

\iclrfinalcopy % Uncomment for camera-ready version, but NOT for submission.
\begin{document}

\maketitle
% Remove "Published as a conference paper at ICLR 2027"
\lhead{}
\begin{abstract}
Foundation-model agents are often modeled as policies over an observed state. In deployed systems, however, a runtime may intervene only after the model has emitted a semantic proposal, making the proposal both an action candidate and a decision-time observation generated by a history-conditioned process. We show that collapsing this structure into a state-only proposal envelope can preserve proposal coverage while destroying certifiability. In a finite robust interface, the viability kernel of the collapsed model is contained in the physical projection of the history-augmented kernel, and the collapse is lossless exactly when every proposal-conditioned collapsed fiber retains a common robust-safe intervention. This gap can be maximal even with constant-size proposal and history alphabets. The same common-action condition yields a dual result: observing the current proposal can restore robust feasibility when it separates latent modes requiring incompatible interventions. We extend these one-step results over time using exact finite beliefs and standard safety and reachability fixed points, separating indefinite operational viability from finite worst-case verified progress. Controlled model-in-the-loop tests reproduce the predicted obstructions when telemetry or effect verification is removed or intervention authority is restricted. Thus, our contribution is not a new fixed-point calculus, but a characterization of when proposal--history correlation at the model--tool boundary is necessary for certification.
\end{abstract}

\section{Introduction}

Foundation-model agents increasingly act through runtimes that mediate external effects.\citep{zhong2026aiharnessengineeringruntime}. At a systems level, we view such a runtime as a regulatory substrate beneath task-level intelligence: its role is not to generate the model's semantic competence, but to preserve the conditions under which that competence can act safely and productively over time. This view follows the classical control-theoretic separation between a system's task behavior and the feedback mechanisms that maintain viable operation under disturbance \citep{cannon1929homeostasis,ashby1960design,wiener1961cybernetics,hellerstein2004feedback}. In this limited functional sense, the harness plays a role analogous to a brainstem for an intelligent system: it mediates action, reacts to operational state, and protects continued operation without replacing higher-level cognition.

A model may propose a tool call, record update, code execution, retry, or termination; only afterward does the runtime decide whether and how to execute the proposal. This order creates an information structure absent from the usual state-feedback abstraction: the proposal is not only an action candidate, but also an observation generated by a history-conditioned model.

A natural simplification is to replace this history dependence with a state-only envelope containing every proposal that may occur at a given physical state. Such an abstraction is sound in the ordinary coverage sense. But is preserving coverage sufficient for certification? We show that it is not. By merging histories that support different proposals or require different interventions, a state-only envelope can force the runtime to defend against proposal--history combinations that were never jointly relevant to a viable execution. The resulting certificate can therefore be strictly smaller even though no proposal has been omitted.

The key object is a \emph{proposal-conditioned common safe intervention}. For a given runtime observation and proposal, all latent modes that remain possible must admit at least one intervention that is both executable and robustly safe. Collapsing history can enlarge this decision fiber until the common intersection disappears. Conversely, because the proposal is observed before intervention, conditioning on it can refine a coarse fiber and restore a common intersection. The same common-action obstruction therefore explains both certification loss under abstraction and certification recovery from proposal information.

We formalize this effect in a finite robust model of the model--tool boundary. Our main result shows that the viability kernel under a coverage-sound state-only history collapse is contained in the physical projection of the history-augmented kernel. We give an exact criterion for losslessness and a constant-alphabet construction in which the entire projected viable region is lost under collapse. Importantly, this strict loss can arise from proposal--history correlation alone, even when authority, disturbances, and physical transitions are history-independent. This isolates the abstraction error from the classical viability machinery used to analyze it.

The proposal itself is also decision-relevant information. Under a fixed sound interface, observing the proposal cannot reduce robust feasibility and can strictly increase it when the proposal separates latent modes with incompatible safe-intervention requirements. To extend these one-step results over time, we construct an exact finite information state and apply standard safety and reachability fixed points. This yields two distinct certificates: one for indefinite operational viability and another for verified goal completion within a finite worst-case horizon. The distinction matters because a runtime may remain safe by stalling without certifying progress.

We make three contributions. First, we identify proposal--history correlation as certification-relevant and show that a coverage-sound state-only collapse can strictly shrink the viable region. We characterize losslessness by a proposal-conditioned common-safe-intervention condition and give a constant-alphabet construction exhibiting maximal loss. Second, we establish the dual refinement result: observing the proposal can restore robust feasibility exactly when it resolves an incompatible safe-action intersection. We then extend the interface over time using exact finite beliefs, separating indefinite viability from finite worst-case verified progress. Third, controlled foundation-model-in-the-loop ablations of telemetry, productive authority, and effect verification reproduce the predicted common-action obstructions. Detailed proofs, secondary information bounds, and full evaluation protocols are deferred to the appendices.
\begin{figure*}[t]
    \centering
    \includegraphics[width=\textwidth]{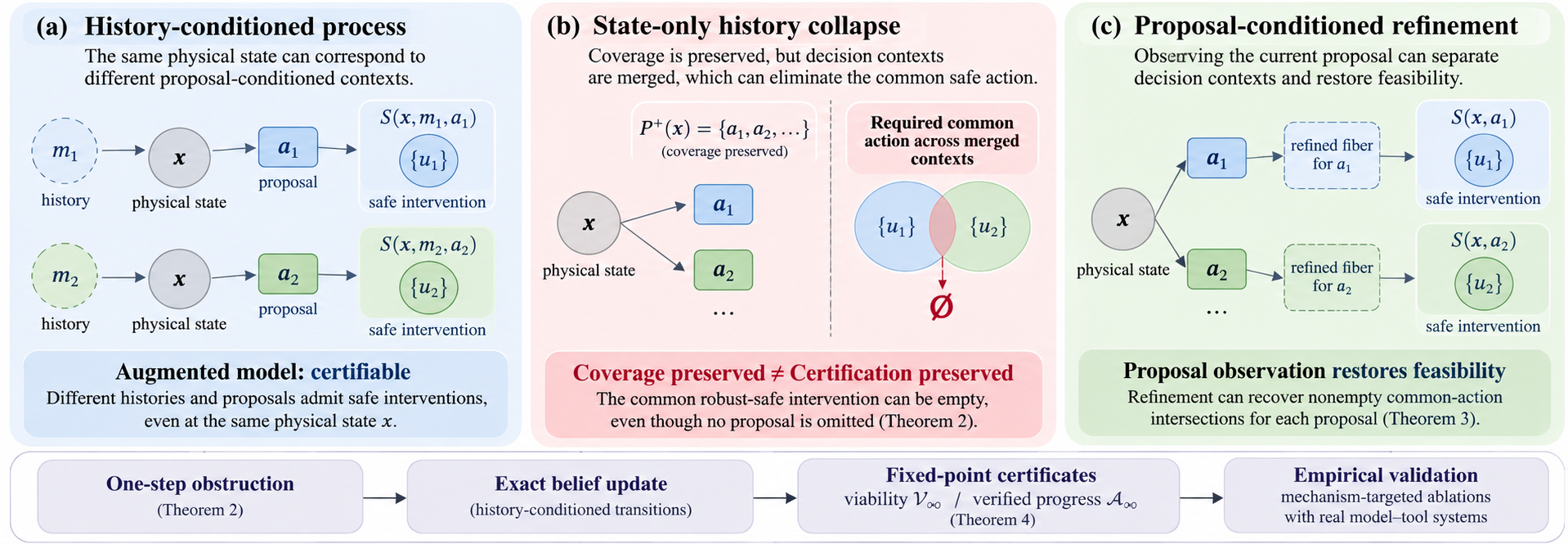}
    \caption{
    \textbf{Core mechanism.}
    History-conditioned proposal contexts can each admit a safe intervention
    (left), while a coverage-sound state-only collapse can merge them and
    eliminate their common robust-safe intervention (center; Theorem~1).
    Observing the current proposal can refine the merged decision context and
    restore robust feasibility (right; Theorem~2).
    }
    \label{fig:core-idea}
\end{figure*}

\section{The proposal-conditioned runtime contract}

We study the boundary at which a model proposal may become an external effect. A wrapper, tool router, monitor/controller pair, or execution stack may implement this boundary; here, \emph{harness} refers to the mediation function rather than to a particular architecture.\citep{zhong2026aiharnessengineeringruntime}. Certification depends on six coupled elements: proposal support, decision-time information, executable authority, the disturbance/transition envelope, verification signals, and causal order.

This contract has a classical control-theoretic interpretation. Requisite-variety arguments emphasize that regulation requires distinctions commensurate with the disturbances and responses that matter to control \citep{ashby1958requisite}; the good-regulator and internal-model traditions emphasize the dependence of effective regulation on an adequate model of the controlled process \citep{conant1970regulator,francis1976internal}. Feedback control and autonomic-computing architectures carry the same regulatory perspective into engineered computing systems \citep{kephart2003autonomic,white2004architectural,hellerstein2004feedback}. Our use of these ideas is structural rather than metaphorical: the harness can certify only distinctions that its interface preserves, and it can realize only interventions that its authority exposes.

The order is
\[
\text{proposal} \rightarrow \text{runtime observation} \rightarrow \text{intervention}
\rightarrow \text{disturbance} \rightarrow \text{transition} \rightarrow \text{verification}.
\]
This order is part of the model. The harness may condition on the proposal and recorded telemetry, but not on a disturbance that occurs later. Likewise, a progress certificate may use an execution outcome only if that outcome is represented in the runtime information state. Thus, a transport acknowledgment does not by itself constitute semantic verification. Appendix A.1 states the complete axioms.

The classical literature motivates the regulatory role; the downstream synthesis machinery is likewise well established: viability theory \citep{aubin2011viability}, supervisory control and imperfect-information games \citep{ramadge1987supervisory,chatterjee2007omega}, belief-state methods \citep{kaelbling1998planning}, and shielding/runtime assurance \citep{konighofer2017shield,alshiekh2018safe,carr2023safe,hobbs2023runtime}. Recent LLM-agent systems likewise enforce runtime rules \citep{wang2026agentspec,doshi2026verifiably,ng2026runtime}. Our question precedes synthesis: \emph{which proposal-conditioned distinctions must survive abstraction for these methods to certify the runtime?} Theorems 1--2 answer this question through the common-action criterion. Thus, the classical control perspective supplies the design principle, while our contribution identifies a foundation-model-specific information condition required to certify it; the novelty does not lie in the fixed-point machinery itself.

\section{Proposal-conditioned feasibility}

We now instantiate the axioms as a finite robust transition interface. The finite model makes proposal support, information, authority, and quantifier order explicit. Extensions to continuous or Borel spaces require additional measurability and compactness assumptions that we do not make here.

Sections 3--6 use this finite robust interface; Appendix A.1 collects the complete proof-domain assumptions. Proposal supports and executable disturbance sets are nonempty; goal membership is verifiable from the runtime information state; and any history relevant to future proposals, authority, observations, disturbances, or transitions is either retained or soundly over-approximated. All guarantees are relative to this model.

At step $t$ the foundation-model policy $\pi_{\theta}\left( \cdot \mid h_{t},z_{t} \right)$ proposes $a_{t}$. We represent its possible outputs at state $x$ by a nonempty support $P(x) \subseteq A$. This may be an over-approximation of outputs across task histories consistent with $x$. The harness observes $y_{t} = O\left( x_{t} \right)$ and the proposal $a_{t}$, and then selects $u_{t} = \mu\left( y_{t},a_{t} \right)$ from its authorized intervention set. Let $U(y,a)$ be the interventions exposed by the interface and $U(x,a) \subseteq U\left( O(x),a \right)$ those actually executable at state $x$. The target is available to the harness only insofar as it is represented in $O(x)$ or the retained history; any unobserved target component must therefore remain in the harness's belief. After the intervention is selected, a tool or environment disturbance $d_{t}$ occurs, and the agent's joint operating and task state evolves according to

\begin{equation}
x_{t + 1} = F\left( x_{t},a_{t},u_{t},d_{t} \right),\quad\quad a_{t} \in P\left( x_{t} \right),\quad d_{t} \in D\left( x_{t},a_{t},u_{t} \right).
\tag{1}\label{eq:runtime-dynamics}
\end{equation}

This order defines the proposal-to-effect gap: the harness may condition on the proposal, but not on an unobserved contemporaneous disturbance. The set $U(x,a)$ contains only interventions that are actually executable; these may include accepting, rewriting, delaying, aborting, retrying, switching tools, or restoring resources. When available, the identity intervention $u=\mathsf{accept}$ recovers the unregulated system.

We distinguish policy classes only when their information restrictions affect the certificate. An \textbf{open-loop} schedule is fixed by the initial condition and clock; a \textbf{static guard} may inspect the current proposal; and a \textbf{state-feedback} or information-state harness may additionally condition on runtime observations or retained history. All impossibility statements below are relative to the explicitly stated information class.

\textbf{Definition 1 (robust operational viability).} A policy makes $C \subseteq K$ invariant if, for every $x_{0} \in C$, every compatible sequence of proposals and disturbances under \eqref{eq:runtime-dynamics} remains in $C$ for all $t \geq 0$. A \emph{sustained} policy additionally reaches $G$ almost surely with finite expected hitting time under a stated stochastic environment, or satisfies a separately stated worst-case progress criterion. Invariance and progress are distinct obligations.

Definition 1 supplies the full-information baseline. The first question is whether a simpler state-only interface preserves that certificate when the model proposal process is history-conditioned. Section 4 shows that it need not.

\section{History collapse can destroy certification}

A foundation model may produce different proposals at the same physical tool state because its proposal process is history-conditioned. Collapsing those histories into a state-only proposal envelope reduces the model while preserving proposal coverage. Theorem 1 shows the cost of this simplification: coverage soundness need not preserve certification. The collapse can create proposal-conditioned decision contexts that the runtime must defend against even though they were never jointly relevant in the augmented process.

For $C \subseteq K$, define the proposal-conditioned robust predecessor

\begin{equation}
\operatorname{Pre}(C) = \{ x \in K:\ \forall a \in P(x)\ \exists u \in U(x,a)\ \forall d \in D(x,a,u),\ F(x,a,u,d) \in C\}.
\tag{2}\label{eq:robust-predecessor}
\end{equation}

The order $\forall a\,\exists u\,\forall d$ follows directly from the interface: the model proposes, the harness intervenes after observing the proposal, and the disturbance follows. Changing this order changes the game. Starting from $C_{0}=K$, let $C_{n+1}=C_n\cap\operatorname{Pre}(C_n)$. Because the model is finite, the sequence stabilizes at $C_{\infty}$. This is the standard guaranteed-viability construction \citep[Secs.~2.9 and 2.11]{aubin2011viability}, applied to the proposal-conditioned interface.

In the finite model, the standard predecessor argument makes $C_{\infty}$ the greatest subset of $K$ admitting a stationary state-and-proposal-dependent policy that is invariant against all modeled proposals and disturbances. Appendix A gives the proof and scope conditions. The key question is what happens when the history-conditioned proposal process is collapsed before this standard synthesis is applied.

\paragraph{History-conditioned proposals and state-only collapse.}
 Let $M$ be a finite Markov-sufficient history abstraction with nonempty admissible fibers $M_x$, and let $\widetilde{P}(x,m)$ be the exact proposal support at augmented state $(x,m)$. The augmented interface has effective authority $\widetilde{U}((x,m),a)$, disturbance set $\widetilde{D}((x,m),a,u)$, and transition

\begin{equation*}
\widetilde{F}\left( (x,m),a,u,d \right) = \left( F^{m}(x,a,u,d),\eta(m,x,a,u,d) \right),
\end{equation*}

where

\begin{equation*}
\eta(m,x,a,u,d) \in M_{F^{m}(x,a,u,d)}.
\end{equation*}

Its viable set is

\begin{equation*}
\widetilde{K} = \{(x,m):x \in K,\ m \in M_{x}\}.
\end{equation*}

Let $\widetilde{C}_{\infty}$ denote the corresponding full-information viability kernel.

To compare this history-conditioned interface with a state-only abstraction, define

\begin{equation*}
P^{+}(x) = \bigcup_{m \in M_{x}}\widetilde{P}(x,m),\quad\quad M_{x,a} = \{ m \in M_{x}:a \in \widetilde{P}(x,m)\},
\end{equation*}

\begin{equation*}
U^{\cap}(x,a) = \bigcap_{m \in M_{x,a}}\widetilde{U}\left( (x,m),a \right),
\end{equation*}

and

\begin{equation*}
\Phi^{+}(x,a,u) = \{ F^{m}(x,a,u,d):m \in M_{x,a},\ d \in \widetilde{D}\left( (x,m),a,u \right)\}.
\end{equation*}

We call $(P^{+},U^{\cap},\Phi^{+})$ the state-only history collapse. Its predecessor is

\begin{equation*}
\operatorname{Pre}^{+}(C) = \left\{ x \in K:\forall a \in P^{+}(x)\ \exists u \in U^{\cap}(x,a)\ \Phi^{+}(x,a,u) \subseteq C \right\},
\end{equation*}

where an empty $U^{\cap}(x,a)$ makes the existential condition false. Let $C_{\infty}^{+}$ be the greatest fixed point obtained from this predecessor.

\textbf{Theorem 1 (History-Collapse Dominance).} If the augmented model is closed under the memory update $\eta$, then

\begin{equation*}
\{(x,m):x \in C_{\infty}^{+},\ m \in M_{x}\} \subseteq {\widetilde{C}}_{\infty},
\end{equation*}

and hence

\begin{equation*}
\boxed{C_{\infty}^{+} \subseteq \operatorname{proj}_{X}\left( {\widetilde{C}}_{\infty} \right).}
\end{equation*}

The inclusion can be strict.

No proposal has been omitted, yet the collapsed runtime must defend against additional proposal-conditioned decision contexts. This is the certification loss caused by discarding proposal--history correlation.

To expose the exact obstruction, let $B = \operatorname{proj}_{X}\left(\widetilde{C}_{\infty}\right)$. For each history mode $m$ supporting proposal $a$ at physical state $x$, define the interventions that keep the physical successor inside $B$,

\begin{equation*}
{\widetilde{S}}_{B}(x,m,a) = \{ u \in \widetilde{U}\left( (x,m),a \right):\forall d \in \widetilde{D}\left( (x,m),a,u \right),\ F^{m}(x,a,u,d) \in B\}.
\end{equation*}

The collapsed proposal-conditioned safe set is exactly the intersection over all histories that the collapse makes compatible with $(x,a)$:

\begin{equation*}
S_{B}^{+}(x,a) = \bigcap_{m \in M_{x,a}}{\widetilde{S}}_{B}(x,m,a).
\end{equation*}

Thus history collapse is lossless on $B$ exactly when every collapsed proposal-conditioned fiber retains a common robust-safe intervention. This common-action intersection is the operational test for certificate preservation.

The intersection can fail because of incompatible proposal support, effective authority, or successor requirements. Even when authority, disturbances, and physical transitions are history-independent, proposal--history correlation alone can make the inclusion strict.

The loss can be maximal: Appendix A gives a construction with two history symbols and two proposal symbols in which the entire projected viable region is lost and the collapsed kernel is empty.

\textbf{Corollary 1 (Exactness Criterion for History Collapse).} Under the assumptions of Theorem 1,

\begin{equation*}
\boxed{C_{\infty}^{+} = B\quad \Leftrightarrow \quad B \subseteq \operatorname{Pre}^{+}(B).}
\end{equation*}

The criterion is local to the final viability region: omitted history may change the interface outside $B$ without changing the certificate. Appendix A.1 gives a stronger global history-rectangularity condition and a finite witness in which strictness is caused solely by replacing $\widetilde{P}(x,m)$ with the state-only union $P^{+}(x)$.

The comparison is between full-information kernels on $(x,m)$ and $x$. If $m$ is not available to the deployed harness, $\widetilde{C}_{\infty}$ is only an upper bound; Section 5 supplies the implementable information-state condition. This leads to the dual question: can information available before intervention recover the distinction lost under collapse?

\section{Proposal observation can restore feasibility}

The history-collapse theorem identifies what is lost when proposal-conditioned distinctions are merged. The current proposal provides the dual perspective: because it is available before intervention, it can refine the runtime's uncertainty about the latent operating mode. The relevant question is not merely whether finer information can help, but which proposal-induced refinements restore a common action lost under the coarse interface.

\begin{equation}
S_{C}(x,a) = \{ u \in U(x,a):\ \forall d \in D(x,a,u),\ F(x,a,u,d) \in C\}.
\tag{3}\label{eq:safe-action-set}
\end{equation}

Let $C_{y,a} = \{ x \in C:O(x) = y,\ a \in P(x)\}$. The proposal may itself reveal information about the state; hence the fiber is restricted by $a \in P(x)$. A memoryless harness cannot choose different interventions for two states in the same fiber. Its policy needs to be defined only on observation-proposal pairs with nonempty fibers.

The same intersection criterion links collapse and refinement: collapse can enlarge a fiber until its common safe action disappears, whereas proposal observation can refine that fiber until a common safe action reappears.

\textbf{Lemma 1 (proposal-conditioned common-action criterion).} There exists a memoryless observation-and-proposal policy $\mu:Y \times A \rightarrow U$ making $C$ robustly invariant if and only if

\begin{equation}
\bigcap_{x \in C_{y,a}}S_{C}(x,a) \neq \varnothing\quad\text{for every }(y,a)\text{ with }C_{y,a} \neq \varnothing.
\tag{4}\label{eq:common-action}
\end{equation}

A single decision rule must therefore be executable and safe for every latent state in the same observation--proposal fiber. Appendix A gives the proof and a distribution-sensitive relaxation.

\subsection{Decision-relevant proposal refinement}

\textbf{Theorem 2 (Decision-Relevant Proposal Refinement).} Fix a candidate invariant region and a sound finite runtime interface. Let $I^{-}$ denote the information available immediately before the current proposal and let observing $A=a$ refine the corresponding latent-state fiber before intervention selection, without changing the underlying proposal support, effective authority, disturbance relation, or transition envelope. A deterministic robust-safe selector using $(I^{-},A)$ exists if and only if every reachable proposal-conditioned refined fiber has a nonempty common robust-safe intervention. Because each refined fiber is a subset of its pre-proposal coarse fiber, this observation refinement cannot reduce robust feasibility. The gain is strict when a reachable coarse fiber has no common robust-safe intervention while its reachable proposal-conditioned subfibers admit such intersections.

Theorem 2 is the constructive dual of Theorem 1. Strict recovery occurs exactly when a coarse fiber lacks a common robust-safe intervention but each reachable proposal-conditioned subfiber admits one. This is a statement about action compatibility, not information volume: mutual information alone does not determine which interventions are jointly safe. Appendix A gives quantitative graph-coloring and Fano-style specializations.

\subsection{Temporal closure: the runtime information-state game}

The preceding results are one-step. Multi-step certification requires a recursive Markov-sufficient information state. Let $\Xi=X\times M$, let $\Lambda(\xi)\subseteq Y\times A$ be the joint output/proposal support, define $\Lambda_{y,a}=\{\xi\in\Xi:(y,a)\in\Lambda(\xi)\}$, and let $B_t^{-}$ be the predicted augmented-state set before observing $(y_t,a_t)$. The exact filter is

\begin{equation}
B_{t} = B_{t}^{-} \cap \Lambda_{y_{t},a_{t}}.
\tag{5}\label{eq:belief-filter}
\end{equation}

Initial-prior and uniform-initialization conditions are stated in Appendix A. For the main recursion, define the finite decision-state space

\[
\mathcal{Q} = \{(B,a):\varnothing \neq B \subseteq \Xi,\ B \subseteq \Lambda_{y,a}\text{ for some }y \in Y\}.
\]

At node $q = (B,a)$, the intervention must be executable at every state still possible:

\[
U_{B}(a) = \bigcap_{\xi \in B}\widetilde{U}(\xi,a).
\]

For $u \in U_{B}(a)$, define the augmented-state successor set and the next decision nodes by

\[
\operatorname{Post}(B,a,u) = \bigcup_{\xi \in B}\{\widetilde{F}(\xi,a,u,d):d \in \widetilde{D}(\xi,a,u)\},
\]

so the prediction step after intervention is

\[
B_{t + 1}^{-} = \operatorname{Post}\left( B_{t},a_{t},u_{t} \right).
\]

The next observed output--proposal pair then filters this predicted set by \eqref{eq:belief-filter}. In the exact finite model, this is the complete belief recursion; under a sound over-approximation, it remains a conservative update.

\[
\operatorname{Next}(B,a,u) = \left\{ (B',a') \in \mathcal{Q}:\ \exists y' \in Y,\ B' = \operatorname{Post}(B,a,u) \cap \Lambda_{y',a'} \neq \varnothing \right\}.
\]

Here, $\operatorname{Post}$ contains every possible augmented successor, and $\operatorname{Next}$ partitions those successors by the next runtime output--proposal pair. A robust selector requires $u\in U_B(a)$ and must keep the entire belief inside $\widetilde K$, rather than only its most likely state. The subset construction is exact for the finite Markov model and conservative for an over-approximated support. In particular, an LLM-generated summary is a valid information state only if it preserves every distinction relevant to future proposal support, authority, observations, and safe transitions. Appendix A gives the exponential worst-case size bound.

\section{Temporal viability and verified progress}

With the information state fixed, standard robust fixed points yield two distinct guarantees: indefinite viability and finite worst-case verified progress.

\subsection{Viability--progress synthesis}

We perform synthesis on the information-state game from Section 5.2.

Let $\widetilde{G} = \{(x,m) \in \Xi:x \in G\}$ be the augmented verified-goal set, and let $\mathcal{Q}_{K} = \{(B,a) \in \mathcal{Q}:B \subseteq \widetilde{K}\}$. Define the full-goal node set $\mathcal{G}_{0} = \{(B,a) \in \mathcal{Q}_{K}:B \subseteq \widetilde{G}\}$. For $\mathcal{C} \subseteq \mathcal{Q}_{K}$, define the robust information-state predecessor

\[
\operatorname{Pre}_{\mathcal{Q}}\left( \mathcal{C} \right) = \left\{ (B,a) \in \mathcal{Q}_{K}:\exists u \in U_{B}(a),\ \operatorname{Post}(B,a,u) \subseteq \widetilde{K},\ \operatorname{Next}(B,a,u) \subseteq \mathcal{C} \right\}.
\]

The greatest fixed point is obtained from

\[
\mathcal{V}_{0} = \mathcal{Q}_{K},\quad\quad\mathcal{V}_{n + 1} = \mathcal{V}_{n} \cap \operatorname{Pre}_{\mathcal{Q}}\left( \mathcal{V}_{n} \right),\quad\quad\mathcal{V}_{\infty} = \nu\mathcal{V}.\left( \mathcal{Q}_{K} \cap \operatorname{Pre}_{\mathcal{Q}}\left( \mathcal{V} \right) \right).
\]

This is the largest set of runtime information states from which an intervention selected using the available information keeps every possible augmented successor and every resulting observation/proposal branch viable. Assume $\mathcal{G}_{0} \subseteq \mathcal{V}_{\infty}$, so every possible post-goal information state admits a safe continuation.

Let $\mathcal{G} = \mathcal{G}_{0}$ be the verified-goal information states. By assumption, $\mathcal{G} \subseteq \mathcal{V}_{\infty}$. Starting from $\mathcal{R}_{0} = \mathcal{G}$, define

\begin{equation}
\begin{aligned}
\mathcal{R}_{n + 1}
&= \mathcal{R}_{n} \cup
\left\{ (B,a) \in \mathcal{V}_{\infty}:
\begin{array}{l}
\exists u \in U_{B}(a),\ \operatorname{Post}(B,a,u) \subseteq \widetilde{K},\\
\operatorname{Next}(B,a,u) \subseteq \mathcal{R}_{n}
\end{array}
\right\},\\
\mathcal{A}_{\infty}
&= \bigcup_{n \geq 0}\mathcal{R}_{n}.
\end{aligned}
\tag{6}\label{eq:progress-attractor}
\end{equation}

\textbf{Theorem 3 (Proposal-Conditioned Viability--Progress Synthesis).}
A decision node lies in $\mathcal{V}_{\infty}$ iff some runtime-information policy keeps every modeled branch in $\widetilde K$ indefinitely. It lies in $\mathcal{R}_{n}$ iff such a policy also reaches a verified goal within at most $n$ transitions on every branch; the least such $n$ is a rank certificate, and a rank-decreasing selector realizes the corresponding finite worst-case completion bound. Appendix A gives the proof.

For $q\in\mathcal{A}_{\infty}$, write $\rho(q):=\min\{n\geq 0:q\in\mathcal{R}_{n}\}$ for this least-rank certificate.

Thus $\mathcal{A}_{\infty} \subseteq \mathcal{V}_{\infty}$. States in $\mathcal{V}_{\infty}\setminus\mathcal{A}_{\infty}$ admit guaranteed continued operation but no finite worst-case verified-completion certificate under the stated model. This does not preclude completion under a stochastic progress law or a richer validated model. The fixed point stabilizes after at most $|\mathcal{Q}_{K}|$ iterations.

Fully observed specializations, stochastic drift conditions, finite-horizon assumption-failure bounds, and belief-complexity results are retained in Appendix A.

\section{Mechanism-targeted evaluation}

The evaluation directly targets the mechanism predicted by the theory: the loss of a common safe productive intervention when decision-time information, authority, or verification is removed. We use deterministic foundation-model-in-the-loop witnesses to realize each obstruction, and random episodes to measure how often the obstructing branches occur.

\subsection{Foundation-model-in-the-loop obstruction tests}

We instantiate the common-action obstruction at a foundation-model proposal boundary while holding the model, semantic tasks, tool schema, backend schedules, and disturbance family fixed. Starting from the full contract, we ablate one resource at a time: decision-time backend telemetry, productive routing authority, or post-execution effect verification. All four contracts remain robustly viable for the specified initial information states, but only the full contract has a finite verified-progress certificate; for $L$ required semantic updates its worst-case rank is at most $2L$.

Each ablation admits a deterministic witness. Without telemetry, two executions that are indistinguishable at decision time disagree on the authoritative backend, leaving only the nonproductive hold action common-safe. With restricted authority, the authoritative backend is known but the required productive route is unavailable. Without effect verification, identical acknowledgments conceal commit versus no-op outcomes, so retrying can violate exactly-once safety while advancing cannot certify completion. We report proposal-envelope adherence, operational violation, physical and verified completion, and intervention count; Appendix B gives the full protocol, confidence intervals, and stratified results.

\paragraph{Results.}
Across the prespecified 1,800 contract-episodes, proposal-envelope adherence was identical across the four contracts, with 428/450 episodes (95.11\%) adhering under each contract. Conditional on adherence, the full harness had zero operational violations, completed and verified all 428 eligible episodes, and satisfied the predicted $T_{\mathrm{exec}}\leq 2L$ bound at every tested task length. The ablations separate the three predicted failure modes: no telemetry produces safe stalling, restricted authority removes guaranteed productive continuation, and no effect verification separates physical completion from certifiable completion. All deterministic obstruction witnesses were reproduced. Table~\ref{tab:contract-level-outcomes} summarizes the contract-level outcomes; Wilson intervals and stratified results are in Appendix B.

\begin{table}[t]
\centering
\caption{Contract-level outcomes conditional on proposal-envelope adherence. Full audits and confidence intervals are in Appendix B.}
\label{tab:contract-level-outcomes}
\begin{tabular}{lccc}
\toprule
\textbf{Contract} & \textbf{Violation} & \textbf{Physical} & \textbf{Verified} \\
\midrule
Full harness & 0/428 & 428/428 & 428/428 \\
No telemetry & 0/428 & 0/428 & 0/428 \\
Restricted authority & 0/428 & 410/428 & 410/428 \\
No effect verification & 0/428 & 410/428 & not certifiable \\
\bottomrule
\end{tabular}
\end{table}

\section{Discussion and conclusion}

A state-only proposal envelope can preserve every model proposal yet destroy certification by erasing proposal--history correlation. Theorem 1 characterizes this loss through the absence of a common robust-safe intervention, and Theorem 2 gives its constructive dual: observing the proposal can restore feasibility when it separates latent modes with incompatible safe actions.

Exact finite beliefs propagate these distinctions over time. The resulting fixed points separate indefinite viability from finite worst-case verified progress, so safe stalling is not a progress certificate. The controlled ablations reproduce the same obstruction when telemetry or effect verification is removed or intervention authority is restricted.

All guarantees are conditional on the stated proposal support, information, authority, disturbance, and verification model. The broader implication is a control-theoretic abstraction principle: a regulatory layer can act only on distinctions preserved by its information interface and only through interventions exposed by its authority. For foundation-model agents, proposal–history correlation is one such certification-relevant distinction. Our results characterize when that distinction must be preserved at the model–tool boundary before standard viability or reachability synthesis is applied.

\subsection*{AI use statement}
In this work, we did not use generative AI tools for research ideation, theoretical development or proofs, experimental design or execution, data generation or analysis, or the formulation of scientific claims and conclusions; other tasks requiring disclosure are not applicable to this work. Additionally, we used generative AI tools solely for language polishing and editorial review of the manuscript, including improving grammar, clarity, presentation, and checking the manuscript for internal consistency. All substantive manuscript content, theoretical results and proofs, experimental methodology and execution, and analysis and interpretation of the results were produced by the authors. We reviewed all AI-assisted edits and independently verified that they preserved the intended technical meaning, mathematical arguments, experimental results, citations, and scientific claims. We take responsibility for the final content of this work, including any text or other material produced with the aid of generative AI.

\bibliography{iclr2027_conference}
\bibliographystyle{iclr2027_conference}

\appendix
\section{Complete theory, proofs, and secondary results}

This appendix collects the technical material compressed in the main narrative: the complete interface axioms, quantitative information consequences, exactness and strictness constructions, distribution-sensitive relaxations, stochastic progress results, assumption-failure bounds, belief-state complexity, and complete proofs.

\subsection{Interface assumptions, initialization, and history-collapse boundary}

\textbf{Axiom A1 (Markov-sufficient augmented substrate).} The runtime state is $\xi_t=(x_t,m_t)$. The physical/task component $x_t$ contains variables relevant to viability and goal verification; $m_t$ retains exactly the history needed for future proposal support, authority, disturbance, transition, and observation relations to be Markov-sufficient. If an exact finite quotient is unavailable, the model must use a sound over-approximation.

\textbf{Axiom A2 (history-conditioned proposal process).} A foundation model generates $a_t \sim \pi_{\theta}(\cdot \mid h_t,z_t)$. On the augmented substrate its supported proposals are $\widetilde{P}(x,m)$. A state-only envelope may use
\[
P^{+}(x)=\bigcup_{m\in M_x}\widetilde{P}(x,m),
\]
but this union preserves coverage while discarding proposal-history correlation.

\textbf{Axiom A3 (interface causality).} The decision order is proposal $\rightarrow$ runtime observation $\rightarrow$ intervention $\rightarrow$ disturbance $\rightarrow$ realized transition $\rightarrow$ verification. In particular, the intervention may depend on the current proposal but not on a contemporaneous disturbance that occurs afterward.

\textbf{Axiom A4 (epistemic boundary).} A deployed harness may condition only on recorded runtime information: telemetry, proposals, past interventions, and verified outputs. It may not condition on hidden $x_t$ or $m_t$ except through an information state consistent with that record.

\textbf{Axiom A5 (effective authority and disturbance closure).} $\widetilde{U}(\xi,a)$ contains only interventions both authorized and physically executable at the decision point. $\widetilde{D}(\xi,a,u)$ contains every disturbance covered by the guarantee, and the augmented transition is closed on the modeled substrate.

\textbf{Axiom A6 (viability-progress conjunction).} Sustained agency in this paper means both preservation of an operational viable set $K$ and satisfaction of an explicit progress condition toward a verified goal $G$. Safety without progress and progress after leaving the stated operating envelope are different claims.

\textbf{Axiom A7 (verification closure).} Any later decision or progress claim that depends on an intervention outcome must use an effect signal included in the runtime information model. Transport success or a parseable tool response is not, by itself, semantic verification.

These axioms make explicit the information pattern that drives the paper. A proposal is unusual because it plays two roles at the same decision point: it is a candidate semantic action and an observation generated by a history-conditioned process. The first role affects what execution is being considered; the second can refine which latent operational modes remain possible.

These axioms have the same scope as the contract summarized in Section 2; they are stated here to keep the main text claim-driven while preserving the full logical boundary of each theorem.

\textbf{Initial-prior and uniform-initialization conditions.}
Using the notation of Section 5.2, let $B_{0}^{-}\subseteq\Xi$ be a nonempty sound prior support before the initial output--proposal pair is observed, so the true initial augmented state $\xi_{0}$ belongs to $B_{0}^{-}$. After observing $(y_{0},a_{0})$, the initial filtered belief is
\[
B_{0}=B_{0}^{-}\cap\Lambda_{y_{0},a_{0}},
\]
which is nonempty for every realized pair consistent with the sound prior. For a specified family $\mathfrak{B}_{0}$ of admissible initial prior supports, define the corresponding initial decision-node family
\[
\mathcal{Q}_{\mathrm{init}}
=
\bigcup_{B^{-}\in\mathfrak{B}_{0}}
\left\{
(B^{-}\cap\Lambda_{y,a},a):
y\in Y,\ a\in A,\ B^{-}\cap\Lambda_{y,a}\neq\varnothing
\right\}.
\]
A uniform robust-viability guarantee over the specified initialization family requires $\mathcal{Q}_{\mathrm{init}}\subseteq\mathcal{V}_{\infty}$; a uniform finite worst-case verified-progress guarantee requires $\mathcal{Q}_{\mathrm{init}}\subseteq\mathcal{A}_{\infty}$. These are initialization conditions on the guarantee, not additional assumptions on later belief updates.

\textbf{Proof of the predecessor characterization in Section 4.} Define $T(C) = C \cap \operatorname{Pre}(C)$. It is monotone under set inclusion. Since $X$ is finite, the descending sequence $C_{0} = K$, $C_{n + 1} = T\left( C_{n} \right)$ stabilizes. At its fixed point $C_{\infty}$, for each $(x,a)$ there is a $u$ whose every allowed successor remains in $C_{\infty}$; selecting such an action defines a stationary invariant policy. If $Q \subseteq K$ is invariant under some stationary full-state policy in the stated policy class, then $Q \subseteq C_{0}$. If $Q \subseteq C_{n}$, that policy's actions show $Q \subseteq \operatorname{Pre}\left( C_{n} \right)$, hence $Q \subseteq C_{n + 1}$. Induction gives $Q \subseteq C_{\infty}$. The finite fixed point is the formal kernel used by the present framework for operational viability; its mathematical source is discrete guaranteed-viability theory \citep{aubin2011viability}, while the proposal-conditioned action order supplies the AI interface semantics. $\square$

\textbf{Proof of Theorem 1.} Let $C_{0}^{+} = K$ and let $C_{n + 1}^{+}$ be the state-only predecessor iteration using $P^{+}$, $U^{\cap}$, and $\Phi^{+}$. Let $\widetilde{K} = \{(x,m):x \in K,\ m \in M_{x}\}$ and let ${\widetilde{C}}_{0} = \widetilde{K}$ with the product predecessor iteration using $\widetilde{P}$, $\widetilde{U}$, $\widetilde{D}$, and $\widetilde{F}$. We prove by induction that

\[
\{(x,m):x \in C_{n}^{+},\ m \in M_{x}\} \subseteq {\widetilde{C}}_{n}.
\]

For $n = 0$ this is the definition of $\widetilde{K}$. Assume it holds at $n$. Take $x \in C_{n + 1}^{+}$ and $m \in M_{x}$. For every $a \in \widetilde{P}(x,m) \subseteq P^{+}(x)$, the state-only predecessor supplies an intervention $u_{x,a} \in U^{\cap}(x,a)$. Hence this action is executable in every product mode, and every product successor has physical projection in $\Phi^{+}\left( x,a,u_{x,a} \right) \subseteq C_{n}^{+}$. Closure of $\eta$ gives a successor memory state $m' \in M_{x'}$ for each such physical successor $x'$. The induction hypothesis therefore gives $(x',m') \in {\widetilde{C}}_{n}$, so $(x,m) \in {\widetilde{C}}_{n + 1}$. Finite-state stabilization proves the first inclusion and its physical projection. This proves dominance; equality need not hold when proposal support depends on retained history. $\square$

\textit{Strictness clause.} Example A.1 below supplies the finite witness for the strict-inclusion clause of Theorem 1.

\textbf{Example A.1} (strictness from proposal--history correlation alone). Let $X = \{ 0,1,\bot\}$, $K = \{ 0,1\}$, and let $M_{x} = \{ L,R\}$ for each $x \in X$, with memory unchanged by every transition. There is one intervention $u$, one disturbance $d$, and the physical transition is independent of memory:

\[
F\left( 0,a_{L},u,d \right) = 0,\quad F\left( 0,a_{R},u,d \right) = \bot,\quad F\left( 1,a_{R},u,d \right) = 1,\quad F\left( 1,a_{L},u,d \right) = \bot,
\]

with $\bot$ absorbing. At both safe physical states, $\widetilde{P}(x,L) = \{ a_{L}\}$ and $\widetilde{P}(x,R) = \{ a_{R}\}$. Hence ${\widetilde{C}}_{\infty} = \{(0,L),(1,R)\}$ and $B = \{ 0,1\}$. But $P^{+}(0) = P^{+}(1) = \{ a_{L},a_{R}\}$, so each physical state fails the state-only predecessor and $C_{\infty}^{+} = \varnothing$. Effective authority, disturbance structure, and physical transitions are independent of history in this construction. The strict loss therefore cannot be attributed to intersecting history-dependent authority or unioning history-dependent successor relations. It arises solely because the state-only union $P^{+}$ destroys the correlation between retained history and proposal support, thereby introducing proposal--physical-state combinations absent from the viable augmented modes. The example concerns the full-information comparison of Theorem 1; partial-observation implementation remains subject to the information-state synthesis of Section 5.2.

\textbf{Proof of Corollary 1.} Write $T^{+}(C) = C \cap \operatorname{Pre}^{+}(C)$, whose greatest fixed point on the finite state space is $C_{\infty}^{+}$. If $C_{\infty}^{+} = B$, fixed-point closure gives $B \subseteq \operatorname{Pre}^{+}(B)$. Conversely, if $B \subseteq \operatorname{Pre}^{+}(B)$, then $B \subseteq K$ and $B \subseteq T^{+}(B)$, so $B$ is post-fixed for the monotone operator $T^{+}$. The greatest fixed point contains every post-fixed set; hence $B \subseteq C_{\infty}^{+}$. Theorem 1 gives $C_{\infty}^{+} \subseteq B$, yielding equality. $\square$

\textbf{Proposition A.1} (history-rectangular sufficient condition). Call the augmented interface history-rectangular on $K$ when, for every $x \in K$, all admissible memory modes have the same proposal support, the same effective authority for each supported proposal, and the same projected successor relation for each proposal and common intervention, and when every successor memory remains in its admissible fiber. Under these conditions,

\[
C_{\infty}^{+} = \operatorname{proj}_{X}\left( {\widetilde{C}}_{\infty} \right).
\]

Proof. Let $\widetilde{\operatorname{Pre}}$ denote the robust predecessor of the augmented interface. For every $C \subseteq K$, history-rectangularity gives the one-step identity

\[
\widetilde{\operatorname{Pre}}\left( \{(x,m):x \in C,\ m \in M_{x}\} \right) = \{(x,m):x \in \operatorname{Pre}^{+}(C),\ m \in M_{x}\}.
\]

Induction from the lifted initial viable set therefore makes every fixed-point iterate coincide under projection, yielding

\[
C_{\infty}^{+} = \operatorname{proj}_{X}\left( {\widetilde{C}}_{\infty} \right).
\]

$\square$

\subsection{Proposal-conditioned information requirements and certificate deficit}

\textbf{Proof of Lemma 1.} If $\mu$ preserves $C$, then for any $(y,a)$ and any $x \in C_{y,a}$, all allowed successors from $\left( x,a,\mu(y,a) \right)$ lie in $C$ and the intervention is executable. Thus $\mu(y,a) \in S_{C}(x,a)$ for every such $x$, proving the nonempty intersection. Conversely, choose $u_{y,a}$ from every nonempty intersection and define $\mu(y,a) = u_{y,a}$ there. For any $x_{t} \in C$, its observation and proposal produce a pair $\left( y_{t},a_{t} \right)$ with $x_{t} \in C_{y_{t},a_{t}}$. The chosen action lies in $S_{C}\left( x_{t},a_{t} \right)$, so every allowed successor is in $C$. Induction proves invariance. $\square$

For a nonempty candidate set $C \subseteq K$, let $A_{C} = \{ a:\exists x \in C,\ a \in P(x)\}$. For each $a \in A_{C}$, form a \textbf{conflict graph} $H_{a}(C)$. Its vertices are states in $C$ at which $a$ is possible; an edge joins $x,x'$ if $S_{C}(x,a) \cap S_{C}(x',a) = \varnothing$. Unsupported proposals are omitted and the chromatic number of an empty graph is defined as zero. Because executability and the safe successor relation are defined by the physical state, this graph can be constructed before choosing an observation map. By Lemma 1, a deterministic observation map paired with a deterministic memoryless harness must assign different labels to adjacent vertices. Consequently,

\begin{equation}
|Y|\ \geq \ \max_{a \in A_{C}}\chi\left( H_{a}(C) \right),
\tag{A.1}\label{eq:conflict-graph-bound}
\end{equation}

where $\chi$ is chromatic number and the maximum ranges only over proposals supported on $C$. Here $|Y|$ counts labels available to the deterministic observation map $O:X \rightarrow Y$; no noisy channel, history, memory, or side information is available to this memoryless selector. For an AI harness, this necessary \textbf{observational variety} bound states how many operational modes requiring incompatible effective interventions the telemetry must distinguish. It is a necessary capacity condition rather than a sufficient controller construction: the full feasibility test remains Eq.~\ref{eq:common-action}, which also checks executability and the common safe-action intersection. If the graph contains a clique of $q$ mutually incompatible modes, at least $q$ distinguishable labels, or $\log_{2}q$ bits in a zero-error code, are required for a memoryless response. Counting labels alone is insufficient: three safe sets $\{ u_{1},u_{2}\}$, $\{ u_{2},u_{3}\}$, and $\{ u_{1},u_{3}\}$ intersect pairwise, so their conflict graph has no edges, yet their common intersection is empty. Equation~\ref{eq:common-action}, not graph coloring, is the full test.

The same obstruction also yields an information bound under noisy sensing. Take $q \geq 2$ equiprobable modes whose safe-action sets are pairwise disjoint, all under the same proposal $a$. Suppose a sensor produces $Y$, the harness selects $U$ using only $Y$ and private randomness independent of the mode conditional on $Y$, and no history or other side information is available. If its robustly unsafe-action probability $\Pr\{ U \notin S_{C}(X,a)\}$ is at most $\varepsilon \leq 1 - 1/q$, then the following bound holds. This event differs from an observed constraint violation: an action lacking a robust guarantee may happen to be safe under one favorable disturbance. A robustly safe action identifies its mode uniquely, so it yields a decoder with error at most $\varepsilon$. Fano's inequality and data processing \citep{cover2006elements} give

\begin{equation}
I(X;Y)\ \geq \ \log_{2}q - h_{2}(\varepsilon) - \varepsilon\log_{2}(q - 1),
\tag{A.2}\label{eq:noisy-information-bound}
\end{equation}

where $h_{2}$ is binary entropy. The bound applies to the specified equiprobable, mutually incompatible modes and a chosen error target. It is not an unconditional entropy law for arbitrary agent harnesses. If history is available, the Markov chain is $X \rightarrow (H,Y) \rightarrow U$ and the same argument lower-bounds $I(X;H,Y)$, not $I(X;Y)$ alone. The bound also highlights an action-order limit: if an adverse event occurs \emph{after} the harness has acted and is not predictable from $Y$, no amount of information about the previous state allows a response tailored to that event; an action must be safe across its entire assumed disturbance set.

\textbf{Proof of the conflict bound \eqref{eq:conflict-graph-bound}.} If $x,x'$ are adjacent in $H_{a}(C)$ and have the same observation label, then they belong to the same fiber $C_{y,a}$, but their safe-action sets are disjoint. This contradicts the common-action condition in equation~\ref{eq:common-action}. Hence $O$ is a proper coloring of each $H_{a}(C)$, requiring at least its chromatic number of labels. The maximum is required because the same sensor must handle every proposal. Pairwise nonconflict is insufficient for global feasibility, as the three-set example above shows. $\square$

\textbf{Proof of the information bound \eqref{eq:noisy-information-bound}.} Each robustly safe action belongs to at most one of the $q$ pairwise disjoint safe-action sets. Define a decoder from the chosen action to that mode, arbitrarily labeling an action safe for none. Whenever $U \in S_{C}(X,a)$, decoding succeeds, so decoder error is at most $\varepsilon$. Fano's inequality gives $H(X \mid U) \leq h_{2}(\varepsilon) + \varepsilon\log_{2}(q - 1)$ over the stated error range. Since $X$ is uniform, $I(X;U) \geq \log_{2}q - h_{2}(\varepsilon) - \varepsilon\log_{2}(q - 1)$. As $X \rightarrow Y \rightarrow U$ is a Markov chain, data processing gives $I(X;Y) \geq I(X;U)$. $\square$

\textbf{Proposition A.2 (Proposal-Conditioned Robust-Certificate Deficit).} Let $\lambda(x,a)$ be any probability distribution supported on pairs with $x \in C$ and $a \in P(x)$. For each observation-proposal fiber with positive mass, define

\[
C_{y,a}^{\lambda} = \{ x \in C_{y,a}:\lambda(x,a) > 0\}.
\]

For any intervention $u \in U$, define the fiber loss

\[
\ell_{y,a}(u) = \sum_{x \in C_{y,a}^{\lambda}}^{}\lambda(x,a)\,\mathbf{1}\{ u \notin S_{C}(x,a)\},
\]

and let

\[
\ell_{y,a}^{*} = \min_{u \in U}\ell_{y,a}(u).
\]

Then

\begin{equation}
L_{C}^{*} = \sum_{(y,a):\sum_{x \in C_{y,a}}^{}\lambda(x,a) > 0}^{}\ell_{y,a}^{*}
\tag{A.3}\label{eq:certificate-deficit}
\end{equation}

is exactly the minimum, over all deterministic memoryless selectors $\mu:Y \times A \rightarrow U$, of

\[
\Pr_{\lambda}\{\mu\left( O(X),A \right) \notin S_{C}(X,A)\}.
\]

Since $S_{C}(x,a) \subseteq U(x,a)$, an intervention that is not executable at the realized state is counted as lacking a robust certificate. Allowing private randomization conditioned only on $(Y,A)$ cannot reduce $L_{C}^{*}$. Moreover,

\[
L_{C}^{*} = 0
\]

if and only if every positive-mass observation-proposal fiber has a common safe action:

\[
\bigcap_{x \in C_{y,a}^{\lambda}}S_{C}(x,a) \neq \varnothing.
\]

If $\lambda$ has full support on every modeled pair $(x,a)$ with $x \in C$ and $a \in P(x)$, this zero-deficit condition is exactly the common-action criterion in equation~\ref{eq:common-action}. For distributions with zero-mass modeled states, $L_{C}^{*}$ is an evaluation-law quantity and does not certify those omitted states.

The common executable set

\[
\mathcal{U}_{y,a}^{\lambda} = \bigcap_{x \in C_{y,a}^{\lambda}}U(x,a)
\]

remains useful as a zero-error authority test. If it is empty, no intervention is executable across the entire fiber; however, this does not imply that the distribution-sensitive certificate deficit of the fiber equals its entire probability mass.

Equation~\ref{eq:certificate-deficit} is the distribution-sensitive counterpart of Lemma 1. It is proposal-conditioned because the loss is computed after conditioning on the model's proposal and telemetry label; a high-capacity model can still have a nonzero deficit when its proposal fibers merge states requiring incompatible repairs. The distribution $\lambda$ defines an evaluation law rather than a worst-case guarantee; a worst-case objective over supported states would be a different criterion.

\textbf{Proof of Proposition A.2.} For a deterministic memoryless selector $\mu:Y \times A \rightarrow U$,

\[
\Pr_{\lambda}\{\mu\left( O(X),A \right) \notin S_{C}(X,A)\} = \sum_{y,a}^{}{\sum_{x \in C_{y,a}^{\lambda}}^{}\lambda}(x,a)\mathbf{1}\{\mu(y,a) \notin S_{C}(x,a)\}.
\]

The choice of $\mu(y,a)$ affects only the corresponding $(y,a)$ fiber, so minimization separates across fibers. Minimizing each inner sum over $u \in U$ gives $\ell_{y,a}^{*}$, and summing the resulting minima gives \eqref{eq:certificate-deficit}. Because $S_{C}(x,a) \subseteq U(x,a)$, any intervention that is not executable at the realized state is automatically counted as uncertified.

If private randomization is allowed, the expected loss on each fiber is a convex combination of the deterministic action losses $\ell_{y,a}(u)$; such a combination cannot be smaller than their minimum. Finally, $\ell_{y,a}^{*} = 0$ exactly when some $u$ belongs to $S_{C}(x,a)$ for every $x$ in the positive-mass support of that fiber, which is equivalent to a nonempty common safe-action intersection. Summing over fibers proves the zero-deficit characterization. $\square$

\textbf{Proof of Theorem 2.} Each proposal-conditioned refined fiber is a subset of its pre-proposal coarse fiber. If a common robust-safe intervention exists on every coarse fiber, the same intervention remains common on every refined subfiber, proving monotonicity under observation refinement. More generally, a refined selector exists exactly when every reachable refined fiber has a nonempty common-safe-action intersection by Lemma 1 applied to the refined information partition. Strict gain occurs when a coarse fiber has empty intersection while the reachable refined subfibers each have nonempty intersections. $\square$

\subsection{Information-state viability and verified progress}

\textbf{Proof of Theorem 3.} Equation~\ref{eq:belief-filter} is the exact set update on the augmented substrate: an augmented state remains possible precisely when it was consistent with the preceding record and the observed output--proposal pair belongs to its joint support $\Lambda(\xi)$. Given a node $(B,a)$ and $u \in U_{B}(a)$, $\operatorname{Post}(B,a,u)$ is the set of all augmented successors of all states in $B$ and all modeled disturbances. For each next pair $(y',a')$, the intersection with $\Lambda_{y',a'}$ is exactly the posterior belief after that pair is observed. Since every successor $\xi'$ has nonempty $\Lambda(\xi')$, the nonempty members of $\operatorname{Next}(B,a,u)$ cover all of $\operatorname{Post}(B,a,u)$. The Markov-sufficiency axiom makes $(B,a)$ a sufficient state for this finite robust game, even when the deployed policy retains the full record.

For safety, define $T\left( \mathcal{C} \right) = \mathcal{Q}_{K} \cap \operatorname{Pre}_{\mathcal{Q}}\left( \mathcal{C} \right)$. It is monotone on finite $\mathcal{Q}_{K}$, so iteration from $\mathcal{Q}_{K}$ stabilizes at its greatest fixed point $\mathcal{V}_{\infty}$. At each node in this fixed point, its predecessor condition supplies one intervention executable at every $\xi \in B$, keeps every augmented successor in $\widetilde{K}$, and sends every possible next information node back into $\mathcal{V}_{\infty}$. Selecting one such intervention for each node therefore preserves the true augmented state in $\widetilde{K}$ indefinitely. Conversely, suppose a policy, possibly depending on its complete record, preserves the true state in $\widetilde{K}$ indefinitely from $q$. For every finite horizon its first intervention must be common-executable on $B$, all successors must lie in $\widetilde{K}$, and every nonempty next information node must admit the corresponding remaining finite-horizon policy. Induction places $q$ in every descending iterate defining $\mathcal{V}_{\infty}$. This proves part 1 and establishes maximality among all policies using the stated runtime information.

For progress, induct on $n$. At $n = 0$, $\mathcal{R}_{0} = \mathcal{G}$ consists exactly of viable decision nodes whose entire belief is in $\widetilde{G}$, so the verified goal is already known. Suppose membership in $\mathcal{R}_{n}$ is equivalent to a policy that preserves $\widetilde{K}$ and reaches a verified-goal node within at most $n$ transitions on every branch. A node newly added at step $n + 1$ has a common executable intervention, all augmented successors remain in $\widetilde{K}$, and every possible next information node belongs to $\mathcal{R}_{n}$; concatenating the witnessing policies reaches the goal within $n + 1$ transitions. Conversely, if a policy guarantees this within $n + 1$ transitions, then either the current node already belongs to $\mathcal{R}_{n}$, or its first intervention sends every possible next information node into $\mathcal{R}_{n}$. That intervention witnesses the predecessor condition. This proves part 2. At each positive least-rank node choose a witnessing intervention. Every possible next node then has strictly smaller rank, so the rank decreases on every branch until a verified goal is reached. Because $\mathcal{G} \subseteq \mathcal{V}_{\infty}$, switching at that point to a viability policy preserves operation after completion. The finite iteration stabilizes after at most $\left| \mathcal{Q}_{K} \right|$ additions, giving a uniform completion bound for every node in $\mathcal{A}_{\infty}$. $\square$

The fully observed relation uses universal proposal sections, not an existential projection. Define the augmented full-state progress recursion by ${\widetilde{R}}_{0} = \widetilde{G}$ and

\[
{\widetilde{R}}_{n + 1} = {\widetilde{R}}_{n} \cup \left\{ \xi \in {\widetilde{C}}_{\infty}:\ \forall a \in \widetilde{P}(\xi)\ \exists u \in \widetilde{U}(\xi,a)\ \forall d \in \widetilde{D}(\xi,a,u),\ \widetilde{F}(\xi,a,u,d) \in {\widetilde{R}}_{n} \right\}.
\]

Define

\[
\widehat{\mathcal{V}} = \{\xi:\forall a \in \widetilde{P}(\xi),\ \left( \{\xi\},a \right) \in \mathcal{V}_{\infty}\},\quad\quad{\widehat{\mathcal{A}}}_{n} = \{\xi:\forall a \in \widetilde{P}(\xi),\ \left( \{\xi\},a \right) \in \mathcal{R}_{n}\}.
\]

When the runtime observation identifies the entire augmented state $\xi$, the interface assumptions agree with the product model in Theorem 1, and verified-goal states have safe continuation, $\widehat{\mathcal{V}} = {\widetilde{C}}_{\infty}$ and ${\widehat{\mathcal{A}}}_{n} = {\widetilde{R}}_{n}$. If no retained history dimension is needed and the state-Markov interfaces coincide, these reduce to $C_{\infty}$ and the history-free progress recursion defined below. The worst-case augmented-state rank is $\max_{a \in \widetilde{P}(\xi)}\rho\left((\{\xi\},a)\right)$. The ordinary set projection of $\mathcal{V}_{\infty}$ or $\mathcal{A}_{\infty}$ is not equivalent: it asks whether at least one current proposal is certified, while the pre-proposal state recursion must cover every supported proposal.

\textbf{History-free fully observed specialization.} When no retained history dimension is needed, suppose the harness observes $x$ and the absorbing goal states admit safe continuation, so $G \subseteq C_{\infty}$. Within the perfect-observation viability kernel, define $R_{0} = G$ and the increasing sequence

\begin{equation}
R_{n + 1} = R_{n} \cup \{ x \in C_{\infty}:\forall a \in P(x)\ \exists u \in U(x,a)\ \forall d \in D(x,a,u),\ F(x,a,u,d) \in R_{n}\}.
\tag{A.4}\label{eq:full-observed-attractor}
\end{equation}

\textbf{Proposition A.3 (Fully Observed Progress Attractor).} A state $x$ belongs to $R_{n}$ if and only if a state-and-proposal-dependent policy can remain in $K$ and force arrival in $G$ within at most $n$ steps against every allowed proposal and disturbance sequence. Thus $R_{\infty} = \bigcup_{n}R_{n}$ is the perfect-observation specialization of $\mathcal{A}_{\infty}$ and has a bounded worst-case completion guarantee. Its induction is in Appendix A.

The rank of $x$ is the least $n$ with $x \in R_{n}$. Selecting at each nonterminal state an intervention that moves every successor to a lower rank provides a concrete progress certificate. A robustly invariant policy may fail this test because it can cycle indefinitely.

\textbf{Fully observed section identity.} Suppose the runtime output identifies the full augmented state $\xi$. Every decision node is then $(\xi,a)$, with $a \in \widetilde{P}(\xi)$. The full-state predecessor is evaluated before the current proposal, so its condition quantifies over every supported $a$. For the safety recursion, induction over its descending iterates shows that all singleton nodes $(\xi,a)$ belong to the information-state iterate exactly when $\xi$ belongs to the corresponding product-state iterate: after an intervention, $\operatorname{Next}$ contains the singleton node for every supported pair at every successor $\xi'$, and the universal section checks all of them. Taking stabilized iterates yields $\widehat{\mathcal{V}} = {\widetilde{C}}_{\infty}$.

For progress, ${\widetilde{R}}_{0} = \widetilde{G}$ agrees with the section ${\widehat{\mathcal{A}}}_{0}$, because verified goal membership is observed and $\widetilde{G} \subseteq {\widetilde{C}}_{\infty}$. If ${\widehat{\mathcal{A}}}_{n} = {\widetilde{R}}_{n}$, then all current proposal nodes lie in $\mathcal{R}_{n + 1}$ exactly when, for every $a \in \widetilde{P}(\xi)$, there is an executable intervention whose every disturbed successor lies in ${\widetilde{R}}_{n}$. This is precisely the displayed recursion for ${\widetilde{R}}_{n + 1}$. Induction gives ${\widehat{\mathcal{A}}}_{n} = {\widetilde{R}}_{n}$ for each $n$. Consequently, the augmented-state worst-case rank is the maximum over current supported proposals of their proposal-conditioned node ranks. A mere existential projection would certify only one favorable proposal and is therefore not the full-state guarantee. $\square$

\textbf{Proof of Proposition A.3.} At $n = 0$, precisely the states already in $G$ can force goal arrival within zero steps. Assume $R_{n}$ is exactly the set forcing arrival within $n$ steps while remaining in $K$. A state outside $R_{n}$ can force arrival within $n + 1$ steps if and only if, for every proposal it might receive, it can choose an intervention whose every disturbance successor lies in $R_{n}$. This is exactly the second term of \eqref{eq:full-observed-attractor}. The action chosen on each such state followed by the inductive policy proves sufficiency; if no such action exists, an adversary can choose a proposal and successor outside $R_{n}$, proving necessity. Induction completes the result. $\square$

\subsection{Probabilistic progress and assumption failures}

For a stochastic model-harness-environment loop, the following additive-drift argument makes the less stringent progress obligation precise.

\textbf{Proposition A.4 (Progress Under a Viable Controller).} Assume the closed loop remains in $K$ almost surely. Let $W:X \rightarrow [0,\infty)$ equal zero on $G$. If, for some $\delta > 0$, every nonterminal history satisfies

\begin{equation}
\mathbb{E}\left[ W\left( x_{t + 1} \right) - W\left( x_{t} \right) \mid \mathcal{F}_{t} \right] \leq - \delta\quad\text{on }\{ t < \tau_{G}\},
\tag{A.5}\label{eq:drift-condition}
\end{equation}

then $\mathbb{E}\left[ \tau_{G} \right] \leq W\left( x_{0} \right)/\delta$. Invariance plus \eqref{eq:drift-condition} gives both continued operating capacity and finite expected completion time. The event $\{ t < \tau_{G}\}$ is in $\mathcal{F}_{t}$ because goal completion is a verified runtime event. Without a progress condition, safety alone does not establish task-directed agency. The proof uses the conditional drift inequality in \eqref{eq:drift-condition}; its application to an AI system requires a defensible potential $W$ and a measured or justified drift bound for the actual model-harness pair.

Robust guarantees also require a valid envelope for model outputs and environmental events. Suppose $x_{0}$ lies in a set already proved invariant whenever the support, sensor, authorization, and disturbance assumptions used in \eqref{eq:runtime-dynamics} hold. Let $E_{t}$ be the $\mathcal{F}_{t + 1}$-measurable event that, at step $t$, at least one of those assumptions fails: a proposal leaves the supported envelope, telemetry is outside its observation model, an intervention is not authorized or executable, a disturbance leaves $D$, or the realized transition is outside $F$. Assume the conditional bound $\Pr\left( E_{t} \mid \mathcal{F}_{t} \right) \leq \epsilon_{t}$ almost surely, where $\epsilon_{t}$ is an externally justified probability bound and not a frequency copied from one finite experiment. Then the union bound yields

\begin{equation}
\Pr\{\exists t < H:x_{t} \notin K\} \leq \sum_{t = 0}^{H - 1}\epsilon_{t}.
\tag{A.6}\label{eq:assumption-failure-bound}
\end{equation}

No independence is needed. The right-hand side should be read as $\min\{ 1,\sum_{t = 0}^{H - 1}\epsilon_{t}\}$; when the sum exceeds one the numerical bound is vacuous. This is a finite-horizon claim; a nonzero per-step miss probability does not by itself furnish an infinite-horizon guarantee. A fixed failure mode that was omitted from the envelope is not automatically an $E_{t}$ event with a known probability; it is a specification failure until a stochastic model is supplied. Conversely, a worst-case viability proof inside a correctly specified envelope is stronger than a small observed failure rate. These two forms of evidence should not be conflated.

\textbf{Proof of Proposition A.4.} Because $\tau_{G}$ is an $\left( \mathcal{F}_{t} \right)$-stopping time, the indicator $\mathbf{1}\{ t < \tau_{G}\}$ is measurable at time $t$. Apply \eqref{eq:drift-condition} to the stopped process and sum the conditional drift inequalities through time $n - 1$:

\[
\mathbb{E}\left[ W\left( x_{n \land \tau_{G}} \right) \right] \leq W\left( x_{0} \right) - \delta\sum_{t = 0}^{n - 1}\Pr\left( \tau_{G} > t \right).
\]

The left side is nonnegative, so $\mathbb{E}\left[ n \land \tau_{G} \right] = \sum_{t < n}^{}\Pr\left( \tau_{G} > t \right) \leq W\left( x_{0} \right)/\delta$. Monotone convergence as $n \rightarrow \infty$ proves the result. $\square$

\textbf{Boundary of guarantee \eqref{eq:assumption-failure-bound}.} Starting in the stated invariant set, on any horizon-$H$ path with no assumption failure $E_{t}$, invariance follows by induction. Therefore an exit event is a subset of $\bigcup_{t < H}E_{t}$. The union bound proves \eqref{eq:assumption-failure-bound}. This bound does not presume independent errors and does not yield an infinite-horizon guarantee from a constant positive $\epsilon_{t}$. $\square$

\subsection{Belief-state construction cost}

\textbf{Proposition A.5 (Belief-State Construction Cost).} Let $N = |\Xi|$ and $d_{\max} = \max_{\xi,a,u}\left| \widetilde{D}(\xi,a,u) \right|$. Under deterministic observations and the standing finite assumptions, exact robust safety synthesis on the augmented game has at most $2^{N} - 1$ nonempty belief states. With explicit state scanning, an explicit subset-predecessor computation therefore has transition-enumeration cost
\begin{equation}
O\!\left(2^{N}N|A||U|d_{\max}\right).
\tag{A.7}\label{eq:belief-complexity}
\end{equation}
Here $d_{\max}$ is the maximum disturbance branching factor; precomputed bitset transitions can remove the explicit $N$ factor under a unit-cost word-operation convention. When proposal generation is state-Markov and no history variable is needed ($M$ is a singleton), the belief-count bound is $2^{|X|}-1$; retaining a validated proposal-history quotient changes it to $2^{|\Xi|}-1$. The exponential dependence is a property of explicit exact belief-space safety construction, not a claim that every implementation must enumerate all beliefs or store them as internal controller states. The belief-count bound is tight: for each $N$, a finite transition interface can make every nonempty subset reachable, as shown in Appendix A. It is neither a lower bound on deployed harness memory nor a complexity claim for stochastic progress synthesis.

\textbf{Proof of Proposition A.5.} Every nonempty filtered belief is a nonempty subset of the $N$-element set $\Xi$, so there are at most $2^{N} - 1$ such beliefs. With successor relations precomputed, a direct implementation scans each candidate belief, proposal, intervention, and allowed disturbance branch; the worst-case scan is $O\left( 2^{N}N|A||U|d_{\max} \right)$, giving \eqref{eq:belief-complexity}. The deterministic observation map filters each generated successor by its label; if all labels are materialized separately, this adds the corresponding output-enumeration factor but does not change the exponential dependence.

To show that the belief-count bound is tight, let the hidden substrate be $\Xi = \{ 1,\ldots,N\}$ with a constant deterministic observation. Every state supports each finite proposal symbol $\sigma_{B}$, indexed by a nonempty subset $B \subseteq \Xi$. Under a common no-op intervention, define the allowed successor set from every current state $i$ after proposal $\sigma_{B}$ to be exactly $B$. Every transition set is nonempty, and from the initial belief $\Xi$, receiving $\sigma_{B}$ and applying the no-op yields exactly the next belief $B$. Hence all $2^{N} - 1$ nonempty beliefs are reachable in this finite interface. This establishes tightness of the belief-space cardinality bound; it does not establish a controller-memory lower bound, since that depends on the harness memory model and what history is available at decision time. $\square$

\section{Complete evaluation protocol}

This appendix follows the logic of the mechanism-targeted evaluation in the main text. We first specify the sampled design and runtime interface, then derive the contract-level predictions and falsification criteria, and finally report deterministic witnesses, conditional runtime outcomes, end-to-end outcomes, and supplementary stratification.

\subsection{Experimental design and evaluation unit}

The sampled main matrix uses three foundation-model families---DeepSeek, Qwen, and GPT-4---on the same 30 semantic tasks (10 each at $L=2,4,6$), four runtime contracts, and five controlled replicate traces per task. Thus $3 \times 30 \times 5 \times 4 = 1{,}800$ contract-episodes are reported after pilot checks, with 450 episodes per contract. The model names are used as family-level identifiers because the evaluation targets the runtime contract rather than a model-version comparison. Within each model--task--replicate cell, the model-side semantic proposal trace is generated once with no backend telemetry in the model context and is then reused across the four contract replays. The corresponding controlled environment trace supplies the authoritative-backend and ambiguous-effect events; different controllers may consume later trace events differently after a retry or safe stall. This paired construction keeps proposal-envelope adherence comparable across contracts while allowing contract-specific execution paths. The controlled tool service fixes the authority, effect, verification, and exactly-once semantics so that the operational contract---not an uncontrolled external API---is the manipulated variable.

The deterministic obstruction witnesses reported below are separate contract checks rather than additional sampled-rate cells. They test the mechanism predicted by the theory; the 1,800 contract-episode matrix is used for the sampled conditional and end-to-end rates.

\subsection{Controlled model-in-the-loop interface}

Each task contains $L$ required semantic updates. The foundation model receives the natural-language task and proposes one semantic operation at a time. Let $R_t$ denote the set of required updates not yet completed. The task-valid proposal envelope is defined by

\[
P_{task}\left( R_{t} \right) = \{ a:\ a\text{ completes exactly one element of }R_{t}\},
\]

with \texttt{finish} admitted only when $R_t=\varnothing$. Thus every proposal inside the formal envelope is productive: a successfully realized valid proposal decreases the task rank $|R_t|$ by one. A schema-invalid operation, an update inconsistent with the task, repetition of an already completed update, or premature \texttt{finish} is recorded separately as a \textbf{proposal-envelope failure}. The robust harness predictions below are conditional on proposals remaining inside $P_{task}$.

Tool execution is mediated by two physical backends,

\[
r_{t} \in \{ A,B\},
\]

exactly one of which is authoritative at each decision point. The backend identity is an operational variable rather than part of the semantic task and is never chosen by the foundation model. The harness receives a semantic model proposal $a_t$ and selects an intervention

\[
u_{t} \in \{\text{route\_A},\text{route\_B},\text{hold}\}.
\]

Routing a proposal to the current authoritative backend permits execution; routing it to the non-authoritative backend is an operational violation. The logical task state is replicated across backend changes, so switching the authoritative endpoint changes the required routing decision but does not erase previously verified task progress. The environment may select either backend before a subsequent decision.

The viable set $K$ requires that no operation be executed on a non-authoritative backend and that every semantic update be committed exactly once. The intervention \texttt{hold} performs no tool execution, preserves $K$, and makes no task progress. A \emph{retry} is not a separate intervention symbol: after a verified no-commit, it means routing the retained semantic proposal again through one of the available routing interventions.

To make effect verification nontrivial, the execution interface also contains a bounded ambiguous-effect disturbance. On the first execution attempt for a semantic update, the tool may either commit the update or leave the logical state unchanged while returning the same transport-level acknowledgment. If authoritative readback reveals that no commit occurred, the retained proposal may be retried and the retry is guaranteed to commit. Re-executing an update that has already committed violates the exactly-once condition. Hence an acknowledgment alone does not distinguish two successor states that require different subsequent actions. Verification is returned as post-execution information and does not consume an additional transition.

The verified goal set is

\[
G = \{ R_{t} = \varnothing\text{ and authoritative readback confirms the target state}\}.
\]

This construction separates semantic proposal generation from operational routing and effect verification: the foundation model determines what task-directed operation to attempt, while the harness determines whether and how that proposal can be executed under the current operational contract.

\subsection{Controlled harness contracts}

Four contracts use the same foundation model, task instances, semantic tool schema, backend schedules, and ambiguous-effect disturbance family.

\textbf{Full harness.} The runtime observes the current authoritative backend, may route to either backend or hold, and receives authoritative post-execution readback. It can therefore select the safe route and distinguish commit from no-commit before deciding whether a retry is required.

\textbf{No telemetry.} The runtime retains both routing interventions and post-execution verification but does not observe which backend is authoritative before routing. Its decision-time information state can therefore contain both $A$ and $B$.

\textbf{Restricted authority.} The runtime observes the authoritative backend and receives verification but exposes only \texttt{route\_A} and \texttt{hold}. States in which $B$ is authoritative are therefore distinguishable but lack the intervention required for productive execution.

\textbf{No effect verification.} The runtime observes the authoritative backend and retains both routing interventions, but receives only the ambiguous transport acknowledgment after execution. It therefore cannot distinguish a committed update from a no-commit outcome before deciding whether to retry the retained proposal or release the next semantic proposal.

No backend telemetry is exposed to the foundation model in any condition. The experimental manipulation is therefore applied at the model-tool execution contract rather than to the model's task prompt.

\paragraph{Paired execution policy and stopping rule.}
For each model--task--replicate cell, proposal-envelope adherence is scored on the shared model-side semantic proposal trace before contract-specific execution. The four runtime controllers then replay that same proposal trace against the corresponding controlled environment trace. Under the full harness, each proposal is routed to the observed authoritative backend; a verified commit releases the next semantic proposal, whereas a verified no-commit causes the retained proposal to be routed again, with the bounded disturbance guaranteeing that this retry commits. Under no telemetry, the robust controller does not guess an authoritative backend: it selects \texttt{hold} at the first productive decision and the episode terminates as a safe non-completion. Under restricted authority, the controller uses \texttt{route\_A} when $A$ is authoritative and otherwise selects \texttt{hold}; the first such stall terminates the episode as a safe non-completion. Under no effect verification, the controller routes each planned proposal once to the observed authoritative backend and never retries after an ambiguous acknowledgment; after the planned semantic proposal trace is exhausted, physical completion is scored from evaluator-side ground truth that is unavailable to the runtime, while verified completion is recorded as not certifiable.

A proposal-envelope failure terminates the sampled episode and is included only in the end-to-end denominator; contract-conditional rates use $E_P$. A safe \texttt{hold} termination is not repeatedly counted as an arbitrary timeout. We write $T_{\mathrm{exec}}$ for the number of routed tool-execution attempts; \texttt{hold} and authoritative readback do not increment this count.

\subsection{Theory-derived predictions and reference values}

For the specified initial information states, the full contract has robust viability and finite verified-progress certificates, with worst-case rank at most $2L$. Each ablation preserves a common safe continuation but removes a common safe productive continuation on a deterministic witness. Formally, $V^{\star}=1$ for all four contracts, whereas $A^{\star}=1$ only for the full contract. The derivation below makes these contract-level reference values explicit.

Let $Q_0$ denote the specified family of initial information states, including both possible authoritative backends. For a harness contract $c$, define

\[
V^{\star}(c) = \mathbf{1}\{ Q_{0} \subseteq \mathcal{V}_{\infty}^{(c)}\},
\]

and

\[
A^{\star}(c) = \mathbf{1}\{ Q_{0} \subseteq \mathcal{A}_{\infty}^{(c)}\}.
\]

Thus $V^{\star}$ means that the contract admits a policy preserving the viable set on every admissible branch from every specified initial information state. $A^{\star}$ additionally means that a policy can force the verified goal in finite worst-case time while preserving viability. These are robust feasibility indicators, not empirical success probabilities.

For the full harness, a task-valid proposal can always be routed to the observed authoritative backend. After execution, authoritative readback distinguishes commit from no-commit. A committed operation advances the task rank; a verified no-commit can be retried safely, and the bounded disturbance guarantees that this retry commits. Consequently,

\[
V^{\star}\left( \text{full} \right) = 1,\quad\quad A^{\star}\left( \text{full} \right) = 1.
\]

Each of the $L$ required semantic updates needs at most two execution transitions, so the initial state has the finite worst-case rank bound

\[
\max_{q \in Q_{0}}\rho(q) = 2L.
\]

The bound is attained when the first attempt of every required update takes the allowed no-commit branch.

Without backend telemetry, a task-valid proposal $a$ has different safe routing actions in the two indistinguishable operational modes:

\[
S(A,a) = \{\text{route\_A},\text{hold}\},\quad\quad S(B,a) = \{\text{route\_B},\text{hold}\}.
\]

Their only common safe intervention is therefore

\[
S(A,a) \cap S(B,a) = \{\text{hold}\}.
\]

The runtime can preserve viability by holding, but no common intervention both preserves $K$ and advances the task. Hence

\[
V^{\star}\left( \text{no telemetry} \right) = 1,\quad\quad A^{\star}\left( \text{no telemetry} \right) = 0.
\]

Under restricted authority, telemetry identifies the authoritative backend, but when $B$ is authoritative the available intervention set contains no safe productive route. Holding again preserves $K$ without reducing task rank. Therefore

\[
V^{\star}\left( \text{restricted authority} \right) = 1,\quad\quad A^{\star}\left( \text{restricted authority} \right) = 0.
\]

Finally, consider the no-effect-verification contract after an ambiguous acknowledgment. The runtime information state contains both a branch in which the proposed update committed and a branch in which it did not. Retrying is productive in the no-commit branch but violates the exactly-once condition in the committed branch; advancing without retry preserves the committed branch but leaves the no-commit branch incomplete. Holding is common-safe but cannot force progress. Thus

\[
V^{\star}\left( \text{no effect verification} \right) = 1,\quad\quad A^{\star}\left( \text{no effect verification} \right) = 0.
\]

The resulting reference values should not be read as empirical rate predictions. In particular, $A^{\star}$ does \textbf{not} predict that an empirical completion rate must be zero. It states that no policy in the specified contract can guarantee safe verified completion on every admissible branch. A favorable random trace may still complete successfully. Conversely, $A^{\star}$ for the full harness is a conditional worst-case statement: it applies only while the model proposals, backend transitions, execution disturbances, and verifier behavior remain inside the stated envelope. Table~\ref{tab:theory-reference-values} summarizes these contract-level predictions.

\begin{table}[p]
\centering

\rotatebox{90}{%
\begin{minipage}{0.92\textheight}
\centering

\end{minipage}%
}
\end{table}

\subsection{Measurements and falsification criteria}

We report proposal-envelope adherence, operational violation, physical completion, verified completion, and intervention count. Contract outcomes are reported both end-to-end and conditional on proposal-envelope adherence.

Let

\[
E_{P} = \{\text{all model proposals required by the episode remain in }P_{task}\}.
\]

For each contract we report both end-to-end performance and the contract-conditional quantities

\[
\Pr\left( \text{violation} \mid E_{P} \right),\quad\quad\Pr\left( \text{physical completion} \mid E_{P} \right),\quad\quad\Pr\left( \text{verified completion} \mid E_{P} \right).
\]

Unconditional outcomes evaluate the complete agent; conditioning on $E_P$ isolates the runtime contract under the formal proposal assumption. Proposal-envelope failure is therefore a model/interface-envelope failure rather than evidence for or against the contract theorem.

Deterministic witness outcomes are contract checks, not fitted targets. Random-episode rates receive confidence intervals and are not substitutes for worst-case certification.

The evaluation therefore keeps the two evidence types distinct: deterministic witnesses test whether the predicted common-action obstruction is realized by the controlled interface, whereas sampled episode rates characterize outcomes under the chosen model, task, and environment-trace population.

The deterministic witnesses are the primary tests. Under the full contract, every in-envelope trace must preserve $K$ and reach the verified goal within $2L$ execution transitions.

\textbf{No telemetry:} pair decision-identical executions with authoritative backend $A$ versus $B$. Any productive route is unsafe in one world; the only common safe action is nonproductive hold.

\textbf{Restricted authority:} set $B$ authoritative while exposing only \texttt{route\_A} and hold. No available intervention is both safe and productive.

\textbf{No effect verification:} pair identical acknowledgments for commit versus no-op. Retry violates exactly-once safety on the committed branch; advancing cannot guarantee completion on the no-op branch.

An in-envelope contradiction that survives implementation and side-channel audit challenges the formal interface mapping. Reproducing a witness provides mechanism-level evidence for the common-action obstruction under the corresponding contract; it does not establish deployment safety or directly test the history-collapse construction. Aggregate rates measure branch prevalence only.

\subsection{Deterministic obstruction tests}

Table~\ref{tab:obstruction-witnesses} reports the mechanism-level witnesses before any aggregate outcome rates are used. The ``No telemetry productive-only'' row is a diagnostic variant of the no-telemetry witness, not a fifth harness contract: it records the consequence of requiring a productive route when the two authoritative-backend states are decision-indistinguishable.

\begin{table}[p]
\centering

\rotatebox{90}{%
\begin{minipage}{0.92\textheight}
\centering

\end{minipage}%
}
\end{table}

\subsection{Conditional runtime results}

Conditional results answer the runtime question under the formal proposal assumption $E_P$. We first audit the full harness because it is the only contract with a finite worst-case verified-progress certificate. Table~\ref{tab:execution-bound-audit} checks the $T_{\mathrm{exec}}\leq 2L$ bound, and Table~\ref{tab:execution-efficiency} reports the corresponding execution counts and retry overhead.

\begin{table}[p]
\centering

\rotatebox{90}{%
\begin{minipage}{0.92\textheight}
\centering

\caption{Theory-derived contract reference values}
\label{tab:theory-reference-values}

\resizebox{\linewidth}{!}{%
\begin{tabular}{lccclll}
\hline
\textbf{Harness contract}
& \textbf{$V^{\star}$}
& \textbf{$A^{\star}$}
& \textbf{Finite worst-case verified-goal bound}
& \textbf{Common-safe continuation}
& \textbf{Common-safe productive continuation}
& \textbf{Formal obstruction} \\
\hline
Full harness
& 1
& 1
& $2L$ transitions
& Yes
& Yes
& None under stated envelope \\

No telemetry
& 1
& 0
& None
& Yes: \texttt{hold}
& No
& Decision-time information \\

Restricted authority
& 1
& 0
& None
& Yes: \texttt{hold}
& No for all $Q_0$
& Effective authority \\

No effect verification
& 1
& 0
& None
& Yes: \texttt{hold}
& No after ambiguous effect
& Post-execution information / verification \\
\hline
\end{tabular}
}

\caption{Deterministic obstruction witnesses}
\label{tab:obstruction-witnesses}

\resizebox{\linewidth}{!}{%
\begin{tabular}{lllll}
\hline
\textbf{Witness} & \textbf{Condition} & \textbf{Common safe action} & \textbf{Productive-action status} & \textbf{Theory result} \\
\hline
No telemetry
& A/B authoritative but indistinguishable
& hold
& None common to both worlds
& No common productive-safe action \\

No telemetry productive-only
& Balanced A/B pair
& None
& Forced \texttt{route\_A} or \texttt{route\_B}; unsafe in one branch
& 50.0\% violation \\

Restricted authority
& B authoritative; \texttt{route\_A} + hold only
& hold
& None
& No productive-safe action under allowed authority \\

No effect verification
& commit/no-op share same acknowledgment
& hold
& None guaranteed-safe + productive
& No guaranteed-safe productive action \\
\hline
\end{tabular}
}

\caption{Full-harness execution-bound audit}
\label{tab:execution-bound-audit}

\resizebox{\linewidth}{!}{%
\begin{tabular}{cccccccc}
\hline
\textbf{Task length L}
& \textbf{Nominal episodes}
& \textbf{$E_P$ eligible}
& \textbf{Violation}
& \textbf{Physical completion}
& \textbf{Verified completion}
& \textbf{Allowed $T_{\mathrm{exec}}$}
& \textbf{Observed $T_{\max}$} \\
\hline
2 & 150 & 143 & 0/143 & 143/143 & 143/143 & $2 \leq T \leq 4$ & 4 \\
4 & 150 & 143 & 0/143 & 143/143 & 143/143 & $4 \leq T \leq 8$ & 8 \\
6 & 150 & 142 & 0/142 & 142/142 & 142/142 & $6 \leq T \leq 12$ & 12 \\
Total & 450 & 428 & 0/428 & 428/428 & 428/428 & --- & 12 \\
\hline
\end{tabular}
}

\vspace{0.6em}

\caption{Observed full-harness execution efficiency. First-attempt commit rates are empirical; the aggregate is 1625/1710 = 95.03\%, not a preset scenario parameter.}
\label{tab:execution-efficiency}

\resizebox{\linewidth}{!}{%
\begin{tabular}{cccccccc}
\hline
\textbf{L}
& \textbf{Eligible episodes}
& \textbf{Required updates}
& \textbf{Observed first-attempt commits}
& \textbf{Verified no-commits / retries}
& \textbf{Total executions}
& \textbf{Mean $T_{\mathrm{exec}}$}
& \textbf{Hard maximum} \\
\hline
2 & 143 & 286 & 272 (95.10\%) & 14 & 300 & 2.10 & $\leq 4$ \\
4 & 143 & 572 & 543 (94.93\%) & 29 & 601 & 4.20 & $\leq 8$ \\
6 & 142 & 852 & 810 (95.07\%) & 42 & 894 & 6.30 & $\leq 12$ \\
Total & 428 & 1710 & 1625 (95.03\%) & 85 & 1795 & 4.19/episode & $\leq 12$ \\
\hline
\end{tabular}
}

\end{minipage}%
}
\end{table}

Table~\ref{tab:contract-task-length} then stratifies the contract-conditional outcomes by task length.

\subsection{Theory-to-experiment falsification audit}

Table~\ref{tab:falsification-audit} summarizes the theory-derived targets against the deterministic ablation witnesses and the full-harness conditional bound check. Here ``PASS'' means that the observed test matched the theory-derived target; it does not mean that an ablated contract achieved verified completion.

\begin{table}[p]
\centering

\rotatebox{90}{%
\begin{minipage}{0.92\textheight}
\centering

\caption{Contract $\times$ task-length stratified outcomes}
\label{tab:contract-task-length}
\resizebox{0.80\linewidth}{!}{%
\begin{tabular}{lccccc}
\hline
\textbf{Contract}
& \textbf{L}
& \textbf{$E_P$ / n}
& \textbf{Violation $|\,E_P$}
& \textbf{Physical $|\,E_P$}
& \textbf{Verified $|\,E_P$} \\
\hline
Full & 2 & 143/150 = 95.33\% & 0/143 & 143/143 = 100\% & 143/143 = 100\% \\
Full & 4 & 143/150 = 95.33\% & 0/143 & 143/143 = 100\% & 143/143 = 100\% \\
Full & 6 & 142/150 = 94.67\% & 0/142 & 142/142 = 100\% & 142/142 = 100\% \\
No telemetry & 2 & 143/150 & 0/143 & 0/143 & 0/143 \\
No telemetry & 4 & 143/150 & 0/143 & 0/143 & 0/143 \\
No telemetry & 6 & 142/150 & 0/142 & 0/142 & 0/142 \\
Restricted authority & 2 & 143/150 & 0/143 & 137/143 = 95.80\% & 137/143 = 95.80\% \\
Restricted authority & 4 & 143/150 & 0/143 & 137/143 = 95.80\% & 137/143 = 95.80\% \\
Restricted authority & 6 & 142/150 & 0/142 & 136/142 = 95.77\% & 136/142 = 95.77\% \\
No effect verification & 2 & 143/150 & 0/143 & 137/143 = 95.80\% & not certifiable \\
No effect verification & 4 & 143/150 & 0/143 & 137/143 = 95.80\% & not certifiable \\
No effect verification & 6 & 142/150 & 0/142 & 136/142 = 95.77\% & not certifiable \\
\hline
\end{tabular}
}

\caption{Theory-derived falsification audit}
\label{tab:falsification-audit}

\resizebox{\linewidth}{!}{%
\begin{tabular}{lllll}
\hline
\textbf{Contract}
& \textbf{Theory target}
& \textbf{Experimental hard test}
& \textbf{Empirical / witness result}
& \textbf{Verdict} \\
\hline

Full harness
& $V^{\star} = 1, A^{\star} = 1$
& All $E_P$-eligible full-harness episodes: zero operational violations, physical and verified completion, and $T_{\mathrm{exec}} \leq 2L$
& 0/428 violations; 428/428 physical and verified completions; $T_{\mathrm{exec}} \leq 2L$ in 428/428
& PASS \\

No telemetry
& $V^{\star} = 1, A^{\star} = 0$
& Balanced A/B indistinguishable witness
& Witness reproduced
& PASS \\

Restricted authority
& $V^{\star} = 1, A^{\star} = 0$
& B-authoritative + \texttt{route\_A}/hold only
& Witness reproduced
& PASS \\

No effect verification
& $V^{\star} = 1, A^{\star} = 0$
& Indistinguishable commit/no-op acknowledgment
& Witness reproduced
& PASS \\

Overall
& Three predicted obstructions
& Deterministic witness reproduction
& 3/3
& PASS \\
\hline
\end{tabular}
}

\caption{Observed end-to-end model-agent outcomes. In tables reporting sampled episode rates, percentages are empirical ratios from the displayed counts}
\label{tab:end-to-end-outcomes}

\resizebox{\linewidth}{!}{%
\begin{tabular}{lccc}
\hline
\textbf{Contract}
& \textbf{End-to-end physical completion}
& \textbf{End-to-end verified completion}
& \textbf{Safe non-completion / residual} \\
\hline

Full harness
& 428/450 = 95.11\%
& 428/450 = 95.11\%
& 22/450 = 4.89\% \\
& [92.71, 96.75]
& [92.71, 96.75]
& \\

No telemetry
& 0/450 = 0.00\%
& 0/450 = 0.00\%
& 450/450 = 100\% safe non-completion \\

Restricted authority
& 410/450 = 91.11\%
& 410/450 = 91.11\%
& 40/450 = 8.89\% \\
& [88.12, 93.40]
& [88.12, 93.40]
& \\

No effect verification
& 410/450 = 91.11\%
& not certifiable
& Physical non-completion: 40/450 = 8.89\% \\
& [88.12, 93.40]
&
& \\
\hline
\end{tabular}
}

\end{minipage}%
}
\end{table}

\subsection{End-to-end agent outcomes}

End-to-end outcomes restore proposal-envelope failures to the denominator and therefore evaluate the complete model--runtime system rather than the runtime contract conditional on $E_P$. Table~\ref{tab:end-to-end-outcomes} reports these unconditional outcomes.

\subsection{Supplementary stratification and uncertainty}

The remaining tables test whether the aggregate pattern is concentrated in a particular model family or task-length cell and provide confidence intervals for the repeated supplementary proportions. In the compact tables, NT denotes no telemetry, RA denotes restricted authority, NV denotes no effect verification, and P/G denotes physical/verified completion.

\begin{table}[p]
\centering

\rotatebox{90}{%
\begin{minipage}{0.92\textheight}
\centering

\caption{Model-family $\times$ contract outcomes}
\label{tab:model-contract-outcomes}

\resizebox{0.75\linewidth}{!}{%
\begin{tabular}{llcccc}
\hline
\textbf{Model family}
& \textbf{Contract}
& \textbf{$E_P$ / n}
& \textbf{Violation}
& \textbf{Physical completion}
& \textbf{Verified completion} \\
\hline
DeepSeek & Full & 143/150 & 0/143 & 143/143 & 143/143 \\
DeepSeek & NT & 143/150 & 0/143 & 0/143 & 0/143 \\
DeepSeek & RA & 143/150 & 0/143 & 137/143 & 137/143 \\
DeepSeek & NV & 143/150 & 0/143 & 137/143 & not certifiable \\
Qwen & Full & 143/150 & 0/143 & 143/143 & 143/143 \\
Qwen & NT & 143/150 & 0/143 & 0/143 & 0/143 \\
Qwen & RA & 143/150 & 0/143 & 137/143 & 137/143 \\
Qwen & NV & 143/150 & 0/143 & 137/143 & not certifiable \\
GPT-4 & Full & 142/150 & 0/142 & 142/142 & 142/142 \\
GPT-4 & NT & 142/150 & 0/142 & 0/142 & 0/142 \\
GPT-4 & RA & 142/150 & 0/142 & 136/142 & 136/142 \\
GPT-4 & NV & 142/150 & 0/142 & 136/142 & not certifiable \\
\hline
\end{tabular}
}

\vspace{0.6em}

\caption{Model family $\times$ task length population cells}
\label{tab:model-task-population}

\resizebox{0.75\linewidth}{!}{%
\begin{tabular}{cccccccc}
\hline
\textbf{Model}
& \textbf{L}
& \textbf{$E_P$ / 50}
& \textbf{Full P/G}
& \textbf{NT P/G}
& \textbf{RA P/G}
& \textbf{NV Physical}
& \textbf{NV Verified} \\
\hline
DeepSeek & 2 & 48/50 = 96\% & 48/48 & 0/48 & 46/48 & 46/48 & not certifiable \\
DeepSeek & 4 & 48/50 = 96\% & 48/48 & 0/48 & 46/48 & 46/48 & not certifiable \\
DeepSeek & 6 & 47/50 = 94\% & 47/47 & 0/47 & 45/47 & 45/47 & not certifiable \\
Qwen & 2 & 48/50 = 96\% & 48/48 & 0/48 & 46/48 & 46/48 & not certifiable \\
Qwen & 4 & 47/50 = 94\% & 47/47 & 0/47 & 45/47 & 45/47 & not certifiable \\
Qwen & 6 & 48/50 = 96\% & 48/48 & 0/48 & 46/48 & 46/48 & not certifiable \\
GPT-4 & 2 & 47/50 = 94\% & 47/47 & 0/47 & 45/47 & 45/47 & not certifiable \\
GPT-4 & 4 & 48/50 = 96\% & 48/48 & 0/48 & 46/48 & 46/48 & not certifiable \\
GPT-4 & 6 & 47/50 = 94\% & 47/47 & 0/47 & 45/47 & 45/47 & not certifiable \\
\hline
\end{tabular}
}

\end{minipage}%
}
\end{table}

\begin{table}[t]
\centering
\rotatebox{90}{%
\begin{minipage}{0.92\textheight}
\centering
\caption{Wilson CI lookup for supplementary cells. Here 95\% denotes the confidence level, not a target outcome rate.}
\label{tab:wilson-ci}
\begin{tabular}{ccc}
\hline
\textbf{Observation} & \textbf{Point estimate} & \textbf{95\% Wilson CI} \\
\hline
48/50 & 96.00\% & [86.54, 98.90] \\
47/50 & 94.00\% & [83.78, 97.94] \\
48/48 & 100.00\% & [92.59, 100] \\
47/47 & 100.00\% & [92.44, 100] \\
0/48 & 0.00\% & [0, 7.41] \\
0/47 & 0.00\% & [0, 7.56] \\
46/48 & 95.83\% & [86.02, 98.85] \\
45/47 & 95.74\% & [85.75, 98.83] \\
\hline
\end{tabular}
\end{minipage}%
}
\end{table}

\end{document}